\documentclass[10pt,twocolumn,letterpaper]{article}

\usepackage[pagenumbers]{cvpr} 

\usepackage{graphicx}
\usepackage{times}
\usepackage{helvet}
\usepackage{courier}
\usepackage{amsmath}
\usepackage{algorithm}
\usepackage{algorithmic}
\usepackage{csquotes} 
\usepackage{color}
\usepackage{paralist}
\usepackage{amssymb}
\usepackage{indentfirst}
\usepackage{float}
\usepackage{multirow}
\usepackage{cite}
\usepackage{mathrsfs}

\usepackage{textcomp,booktabs}
\usepackage{amssymb}
\usepackage{pifont}
\usepackage[misc]{ifsym}

\usepackage{mathrsfs}
\usepackage{color}
\usepackage{xcolor}
\usepackage{tabularx}

\definecolor{citecolor}{HTML}{0071bc} 
\definecolor{SeaGreen4}{RGB}{0,205,102} 
\definecolor{black}{RGB}{106,90,205} 
\definecolor{DarkRed}{RGB}{178,34,34}

\usepackage{color}
\usepackage{xcolor}
\definecolor{citecolor}{HTML}{0071bc} 
\definecolor{SeaGreen4}{RGB}{0,205,102} 
\definecolor{SlateBlue}{RGB}{106,90,205} 
\definecolor{DarkRed}{RGB}{178,34,34} 
\usepackage[colorlinks, linkcolor=red,  anchorcolor=blue, citecolor=citecolor]{hyperref}

\usepackage{colortbl}
\definecolor{mygray}{gray}{.9}
\definecolor{mypink}{rgb}{.99,.91,.95}
\definecolor{mycyan}{cmyk}{.3,0,0,0}

\usepackage{color}
\usepackage{xcolor}
\definecolor{citecolor}{HTML}{0071bc} 
\definecolor{SeaGreen4}{RGB}{0,205,102} 
\definecolor{SlateBlue}{RGB}{106,90,205} 
\definecolor{DarkRed}{RGB}{178,34,34}

\usepackage[capitalize]{cleveref}
\crefname{section}{Sec.}{Secs.}
\Crefname{section}{Section}{Sections}
\Crefname{table}{Table}{Tables}
\crefname{table}{Tab.}{Tabs.}

\def\paperID{9748}      
\def\confName{CVPR}
\def\confYear{2024}

\title{ Structural Information Guided Multimodal Pre-training \\ for Vehicle-centric Perception}

\author{
    Xiao Wang\textsuperscript{\rm 1,2,3}, 
    Wentao Wu\textsuperscript{\rm 1,2,4}, 
    Chenglong Li\textsuperscript{\rm 1,2,4}\thanks{Corresponding author: Chenglong Li (lcl1314@foxmail.com)},  
    Zhicheng Zhao\textsuperscript{\rm 1,2,4}, \\ 
    Zhe Chen\textsuperscript{\rm 5}, 
    Yukai Shi\textsuperscript{\rm 6},
    Jin Tang\textsuperscript{\rm 1,2,3} \\ 
${^1}$ \small{Information Materials and Intelligent Sensing Laboratory of Anhui Province, Anhui University, Hefei 230601, China} \\
${^2}$ \small{Anhui Provincial Key Laboratory of Multimodal Cognitive Computation, Anhui University, Hefei 230601, China} \\
${^3}$ \small{School of Computer Science and Technology, Anhui University, Hefei 230601, China} \\ 
${^4}$ \small{School of Artificial Intelligence, Anhui University, Hefei 230601, China} \\ 
${^5}$ \small{School of Computing, Engineering and Mathematical Sciences, La Trobe University} \\ 
${^6}$ \small{School of Information Engineering, Guangdong University of Technology, Guangzhou, China} 
}

\begin{document}
\maketitle

\begin{abstract}
Understanding vehicles in images is important for various applications such as intelligent transportation and self-driving system. 
Existing vehicle-centric works typically pre-train models on large-scale classification datasets and then fine-tune them for specific downstream tasks. However, they neglect the specific characteristics of vehicle perception in different tasks and might thus lead to sub-optimal performance. To address this issue, we propose a novel vehicle-centric pre-training framework called VehicleMAE, which incorporates the structural information including the spatial structure from vehicle profile information and the semantic structure from informative high-level natural language descriptions for effective masked vehicle appearance reconstruction. To be specific, we explicitly extract the sketch lines of vehicles as a form of the spatial structure to guide vehicle reconstruction. 
The more comprehensive knowledge distilled from the CLIP big model based on the similarity between the paired/unpaired vehicle image-text sample is further taken into consideration to help achieve a better understanding of vehicles.
A large-scale dataset is built to pre-train our model, termed Autobot1M, which contains about 1M vehicle images and 12693 text information. Extensive experiments on four vehicle-based downstream tasks fully validated the effectiveness of our VehicleMAE. 
The source code and pre-trained models will be released at \textcolor{blue}{\url{https://github.com/Event-AHU/VehicleMAE}}.  
\end{abstract}

\section{Introduction}
Vehicles play a very important role in modern real life, such as the private car, public transport bus, trucks, etc. With the development of artificial intelligence, the problem of vehicle-centered perception has attracted more and more attention, especially for autonomous driving, security monitoring, and smart city. Many computer vision problems are proposed for the vehicles, including vehicle detection and tracking~\cite{chadwick2019distant, wang2022beamTrack}, segmentation~\cite{he2022partimagenet}, attribute recognition~\cite{liu2016Veri776, wang2022PARSurvey, jin2023sequencepar}, re-identification~\cite{wang2020attributereid}, fine-grained classification~\cite{krause20133Stanfordcars}, and text-based retrieval~\cite{scribano2021all}. In the early stages of deep learning, these research topics are usually studied in a relatively independent form. To be more specific, they first collect and annotate a small subset, then, train a neural network from scratch or based on a pre-trained backbone network. Then, they conduct the evaluation on the testing subset. Although good performance can be achieved compared with previous ones, however, these problems are far from being solved due to the complicity of the real world.

\begin{figure}
\center
\includegraphics[width=3.3in]{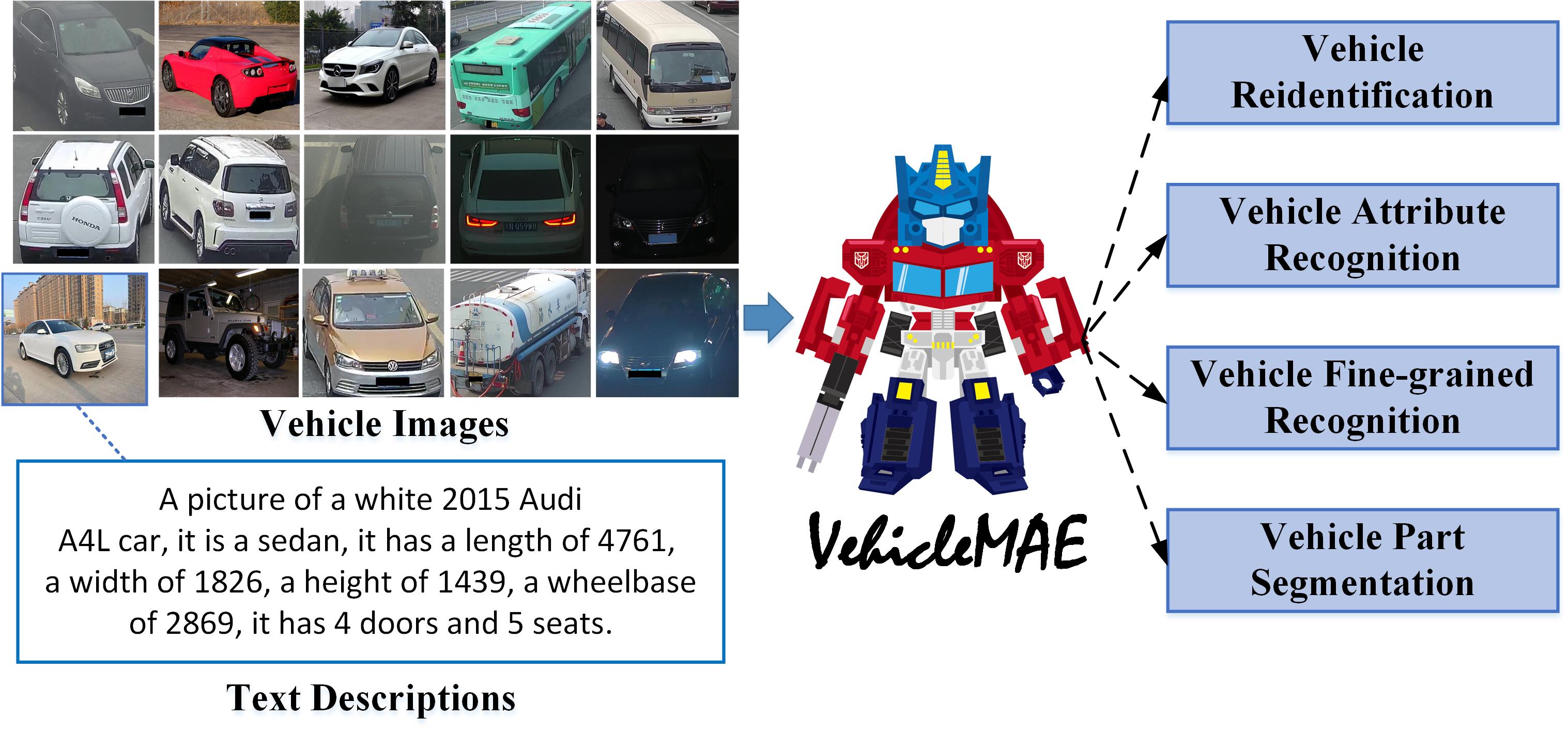}
\caption{Our proposed pre-trained big model VehicleMAE takes the large-scale vehicle images and corresponding natural language descriptions as input and supports multiple downstream vehicle-based tasks.}    
\label{firstIMG}
\end{figure}

Recently, the self-attention-based Transformer~\cite{vaswani2017attention} networks illustrate their powerful ability on capturing long-range relations than the widely used Convolutional Neural Networks (CNN). Also, the big models obtained by stacking the Transformer blocks and pre-training on large-scale datasets demonstrate their amazing performance~\cite{wang2023MMPTMs}. Compared with the traditional small-scale deep neural network, {these} large {Transformer} models have achieved obvious advantages in accuracy and generalization. 
To be specific, Large Language Models (short for LLMs, like the GPT series~\cite{radford2018GPT1, radford2019GPT2, brown2020GPT3, openai2023gpt4}, LLaMA~\cite{touvron2023llama}, T5~\cite{raffel2020LLMMT5}) have demonstrated their powerful performance on various text tasks. They can recognize, summarize, translate, predict, and generate text according to the knowledge learned from massive datasets. 
In addition to the large language models, the pre-training technique also showed superior performance in computer vision. {Some} pioneer models like the ViT~\cite{dosovitskiy2020image} and Swin-Transformer~\cite{liu2021swintransformer} {achieved compelling performance} in the image classification task. These pre-trained models are also widely used in other computer vision tasks, such as object detection, tracking, segmentation, etc. The deep generative models are also significantly boosted with the large model and training data, for example, DALLE~\cite{ramesh2021DALLE}, GPT-4~\cite{openai2023gpt4}, SparkDesk, and ERNIE-Bot. In order to obtain a more comprehensive and accurate model, the researchers pre-train their model on multimodal data~\cite{wang2023MMPTMs}, such as RGB, language, audio, depth, thermal, event stream, etc. Many multimodal {large} models demonstrate that different modalities can achieve the effect of information complementary in the training stage. In other words, it performs better than the unimodal-based model even under modality-missing scenarios in the inference phase.

Inspired by the success of the aforementioned {large} models, in this work, we begin to {investigate} the vehicle-based pre-training algorithms based on MAE (Masked Auto-Encoder)~\cite{he2022mae} for high-performance and generalized vehicle perception. The vanilla MAE takes the high-ratio masked tokens as input and learns to reconstruct them under an auto-encoder framework. Directly adapting the MAE for our vehicle perception is a natural and intuitive approach, however, we believe {that it could be difficult for the original MAE to focus on the vehicle representation without specific designs, and} this framework can be further improved from the following two aspects: 
\textbf{Firstly}, vehicles typically possess certain distinctive visual features, such as their outlines which are often formed by lines and curves, and their consistent coloration. These features are distinct from those of regular objects. We believe that these features can be leveraged during the pre-training phase of the model to enhance its ability to perceive the visual structure and form of vehicles more effectively. 
\textbf{Secondly}, the natural language descriptions of the vehicles can easily be obtained from various vehicle websites. Therefore, from a multimodal pre-training perspective, it is possible to better explore the high-level semantic guidance to improve vehicle reconstruction. Also, existing multimodal big models, such as CLIP~\cite{radford2021learning}, can be fully utilized to further enhance the reconstruction results.

Based on the above observations and reflections, in this paper, we propose a general vehicle-centric pre-training framework that considers both vision and language description for {MAE-based} vehicle perception.
As shown in Fig.~\ref{framework}, our proposed VehicleMAE contains three main modules, including the {MAE}, Structural Prior module, and Semantic Prior module. To be specific, given the input vehicle image, we first partition it into non-overlapping regions. Following the MAE~\cite{he2022mae}, we mask most of the input tokens and feed them into a Transformer encoder for feature representation learning. Then, a Transformer decoder is used for masked token prediction. 
Meanwhile, we also adopt the pre-trained CLIP visual encoder to get the visual representations and the edge detector BDCN network~\cite{he2020bdcn} to get the contour image. For the text input, we adopt the CLIP text encoder to obtain the text embeddings and tune the parameters of our neural network to make the predicted similarity between image-text input consistent with the CLIP model. Note that, the parameters of CLIP are fixed and only the Transformer encoder and decoder are adjustable. 
For the structural prior module, we first partition the contour visual image into non-overlapping patches and fed it into the shared Transformer encoder for contour information extraction. This information can be seen as a guide for the MAE branch for better image reconstruction.

To train our proposed VehicleMAE model, we collect a large-scale dataset that contains about 1M vehicle images, termed \textbf{Autobot1M}. These data are mainly obtained from existing datasets, public visual surveillance systems, and vehicle websites. These data fully reflect the key challenges in vehicle-centric perception, such as illumination, motion blur, viewpoints, and occlusion. Note that, part of these images are crawled from the Internet and the corresponding natural language descriptions (12693 sentences) are also available for pre-training. More details can be found in Section~\ref{autobot1m} and some representative samples of our data can be found in our supplementary materials.

To sum up, the contributions of this paper can be concluded as the following three aspects: 

$\bullet$ We propose the first multimodal pre-training framework for vehicle-centric perception, termed \textbf{VehicleMAE}. The structural contour information and high-level semantic prior are proposed for a more accurate masked token reconstruction. 

$\bullet$ We propose a large-scale dataset to boost the research of pre-training on vehicle images, termed \textbf{Autobot1M}. It contains a total of 1M images and part of them with a corresponding language description. 

$\bullet$ We conduct extensive experiments on four downstream tasks to validate the effectiveness of our proposed VehicleMAE, including Vehicle Attribute Recognition, Vehicle-based Re-identification, Fine-grained Vehicle Classification, and Vehicle Detection.


\section{Related Work} \label{relatedWorks}

In this section, we will review existing works related to our model from the Pre-trained Big Models, Prior Knowledge Guided Pre-training, and Vehicle in Computer Vision. More details can be found in the following survey papers~\cite{wang2023MMPTMs}.

\subsection{Pre-trained Big Models} 
Collecting a large-scale dataset and conducting self-/un-supervised learning on big models is a hot research topic nowadays. The mainstream of the pre-trained models can be divided into the following two categories, i.e., the \emph{contrastive learning} based and \emph{reconstruction based pre-training}. Specifically, the contrastive learning based approach attempt to train the network to discriminate whether the given input pairs are aligned or not. For the single modality based pre-training, the input is usually expanded into Siamese samples using data augmentation for contrastive learning. For example, SimCLR~\cite{chen2020simple} exploits the dataset enhancement and larger batch size in contrastive learning. MoCo~\cite{he2020momentum} is developed based on the idea of momentum encoder and is widely used in many downstream tasks~\cite{wang2021dense}. 
In addition to the simulated or augmented samples for contrastive learning, the multi-modal pre-trained big models are also intuitive and natural to adopt contrastive learning to align different modalities. More in detail, the CLIP~\cite{radford2021learning} and ALIGN~\cite{jia2021scaling} are two big models pre-trained on image-text pairs and achieve amazing performance on the generalized classification task. VideoCLIP~\cite{xu2021videoclip} trains video and text Transformers by comparing temporally overlapping positive video-text pairs with negative sample pairs for nearest-neighbor retrieval. Wang et al. propose the GLIP~\cite{li2022grounded} which unifies the object detection and visual grounding for pre-training.

Reconstruction-based pre-training is another mainstream pre-training technique and representative models like the MAE~\cite{he2022mae} and BERT~\cite{kenton2019bert} are widely used in many applications~\cite{tong2022videomae}. To be specific, BERT~\cite{kenton2019bert} is proposed for natural language processing, which follows the masked token reconstruction framework. MAE (Masked Auto-Encoder) targets reconstructing the original input image from the masked tokens with a high ratio (nearly $75\%$). It is also adapted into the video domain, such as the VideoMAE~\cite{tong2022videomae} and VideoMAE-v2~\cite{wang2023videomae}. MAGE (MAsked Generative Encoder)~\cite{li2023mage} proposed by Li et al. first unifies image generation and self-supervised representation learning. There are also some works that focus on pre-training on human data, such as the HumanBench~\cite{tang2023humanbench}, UniHCP~\cite{ci2023unihcp}, PLIP~\cite{zuo2023plip}. 
Different from existing works for the general object or human-centric based perception, in this work, we propose a pre-trained big model with a focus on vehicles. Our model can empower various downstream tasks and applications related to vehicles, such as vehicle re-identification, vehicle parsing, vehicle retrieval, etc., with significant practical value and potential.

\subsection{Prior Knowledge Guided Pre-training}  
Unlike conventional pre-training solely on raw data, some researchers consider incorporating prior information or domain knowledge into the training of large models~\cite{wu2022wav2clip, kim2022BeamCLIP, liu2023prismer}, aiming to imbue the large models with common sense. One common form of prior knowledge or domain knowledge is the knowledge graph. To be specific, Jaket~\cite{yu2022jaket} is a joint pre-training framework that models the knowledge graph and language simultaneously, as the two modules provide mutually assisted essential information. Liu et al. propose the K-BERT~\cite{liu2020kbert} which is a knowledge graph-assisted language representation model. They build domain knowledge by injecting triples into the sentences, then introduce the soft-position and visible matrix to address the knowledge noise issue. kgTransformer~\cite{liu2022maskReason} is proposed by Liu et al., which attempts to learn high-quality knowledge graph embeddings and adopt Mixture-of-Experts (MoE) sparse activation for further enhancement. PEKG~\cite{wong2023pretrainedKG} is short for pre-trained E-commerce knowledge graph, which can learn the representation of product from EKG. Knowledge-CLIP~\cite{pan2022KGLIPretrain} injects semantic information into the CLIP model for knowledge-based pre-training.

Some researchers also believe that current large models have acquired substantial knowledge, which can be leveraged for iterative development and expansion. For instance, based on pre-trained big models, such as CLIP~\cite{radford2021learning}, BERT~\cite{kenton2019bert}, SAM~\cite{kirillov2023SAM}, and ChatGPT\footnote{\url{https://openai.com/blog/chatgpt}}, a series of derivative large-scale model works have emerged. For example, the authors of Wav2CLIP~\cite{wu2022wav2clip} introduce CLIP~\cite{radford2021learning} for knowledge distillation of audio encoders. They project audio into a shared embedding space with images and text. Kim et al. propose the BeamCLIP~\cite{kim2022BeamCLIP} to transfer the representations of CLIP-ViT into a small target model effectively. Prismer~\cite{liu2023prismer} mines expert knowledge from a wide range of domains and achieves good performance on multiple downstream vision-language tasks. 
Inspired by these works, in this work, we introduce the pre-trained CLIP model and sketch maps to achieve semantic-aware and structure-aware masked token reconstruction.

\subsection{Vehicle in Computer Vision} 
Over the past few years, research on vehicle-centric intelligent visual perception and decision-making problems has never ceased. Among them, the most widely studied visual tasks include vehicle attribute recognition~\cite{liu2016Veri776}, vehicle re-identification~\cite{he2021transreid}, vehicle parts segmentation~\cite{zheng2021rethinking}, and so on. 

For the vehicle attribute recognition task, to be specific, Zhao et al.~\cite{zhao2018grouping} proposed GRL to better exploit potential dependencies between attributes by considering attribute mutual exclusion within clusters and attribute association between clusters. Sarafiano et al.~\cite{sarafianos2017curriculum} combined multi-task learning and course learning to propose a faster-converging framework for visual attribute classification. Cheng et al.~\cite{cheng2022VTB} proposed a framework based on visual language fusion, and fusing bimodality through fine-tuning.

For the vehicle re-identification task, Zhang et al.~\cite{zhang2020part} proposed a partially guided attention network to localise partial regions and combine global and local information, solving the problem that previous work only focused on global appearance features of vehicles. Luo et al.~\cite{luo2019bag} proposed a BNNeck structure to solve the problem that ID loss and triplet loss could not converge simultaneously. Zhuang et al.~\cite{zhuang2020rethinking} proposed a camera-based BN layer to normalise the image features under each camera separately, thus solving the problem of domain differences under multiple cameras. He et al.~\cite{he2021transreid} proposed a pure transform-based re-identification framework to solve the problem that CNN-based models cannot extract global information and fine-grained features better. 

For the vehicle parts segmentation task, Zheng et al.~\cite{zheng2021rethinking} proposed the SETR algorithm, which used VIT instead of CNN as the backbone for feature extraction, and achieved good performance. Xie et al.~\cite{xie2021segformer} proposed a hierarchical transform structure and a lightweight MLP decoding module to solve the problems of inflexible location coding and low computational efficiency. Wan et al.~\cite{wan2023seaformer} propose improved attention through axis compression with the addition of location information, reducing the problems of excessive computation and memory consumption for better deployment on mobile devices. 
Different from previous works which attempt to address one major task, our proposed pre-trained big model VehicleMAE can support multiple downstream tasks and achieves state-of-the-art experimental results.

\begin{figure*}[!htp]
\center
\includegraphics[width=7in]{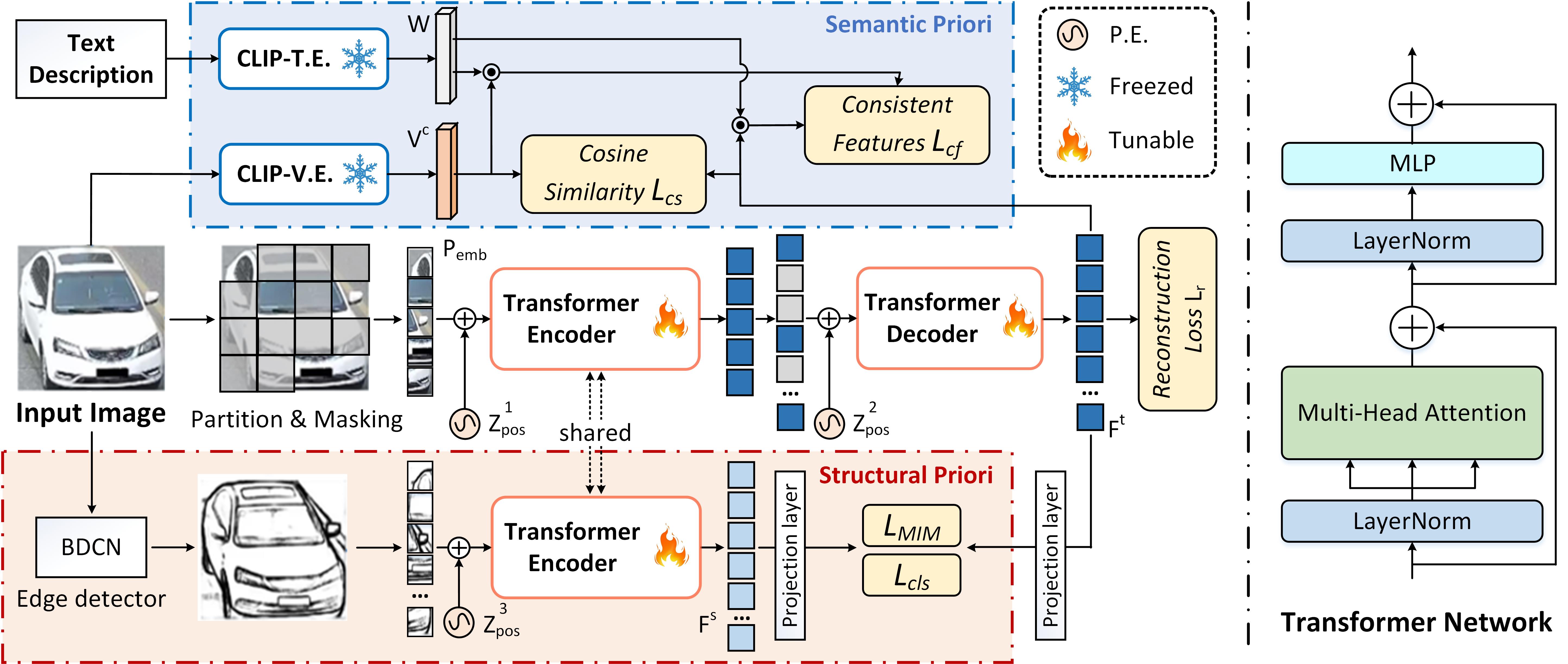}
\caption{ \textbf{An overview of our proposed Structural and Semantic Prior Guided Masked Auto-Encoder Framework for General Vehicle-centric Perception, termed VehicleMAE.} Specifically, we partition the given input vehicle image into non-overlapping tokens, then, mask these tokens with a high ratio. The visible tokens are fed into the Transformer encoder for feature learning, then, the masked tokens are randomly initialized and fed into the Transformer decoder for invisible token prediction. More importantly, we propose the structural prior to guide the image reconstruction procedure using extracted vehicle sketch maps. The CLIP model is also introduced to learn the semantic-aware representations using contrastive learning. Extensive experiments on multiple downstream vehicle-based tasks fully validated the effectiveness of our proposed VehicleMAE. 
} 
\label{framework}
\end{figure*}

\section{Methodology}  \label{method}

In this section, we will first give an overview of our proposed VehicleMAE framework. Then, we will introduce the detailed network architecture, with a focus on the masked auto-encoder, the semantic prior, and the structural prior. After that, we introduce the downstream tasks to validate the effectiveness and generalization of our proposed VehicleMAE.

\subsection{Overview} 
As shown in Fig.~\ref{framework}, our proposed VehicleMAE follows the masked auto-encoder framework which takes the high-ratio masked tokens as the input. Specifically speaking, we first partition the input vehicle image into non-overlapping patches. Most of these tokens are randomly masked, and the remaining ones are used as input to the network. In this paper, we adopt the Transformer encoder and Transformer decoder to achieve the masked token reconstruction. More importantly, we introduce two kinds of prior information to guide the token reconstruction, i.e., the structural and the semantic prior. 
For the structural prior, we adopt the BDCN~\cite{he2020bdcn} edge detector to get the contour map and partition it into non-overlapping regions. The contour representation can be obtained using the shared Transformer encoder, which is an important clue for efficient vehicle reconstruction. 
For the semantic prior, we adopt the off-the-shelf pre-trained vision-language models (CLIP~\cite{radford2021learning} is adopted in this work) to encode the natural language descriptions and build contrastive learning schemes for high-level semantic information guided reconstruction. 
More in detail, the cosine similarity between the CLIP~\cite{radford2021learning} visual embedding and Transformer encoder, and the KL-Distance between the similarity of language embeddings and visual features predicted by the CLIP~\cite{radford2021learning} model and learned Transformer encoder are considered. 
Extensive experiments on multiple downstream tasks demonstrate that the semantic and structural prior improves the pre-training significantly.

\subsection{Network Architecture} 
Our proposed VehicleMAE contains three main modules, i.e., the masked auto-encoder, structural prior, and semantic prior module. 

\noindent
\textbf{Masked Auto-Encoder. } 
Given the vehicle image $I \in \mathbb{R}^{224 \times 224 \times 3}$, we first partition it into 196 non-overlapping patches $P_i \in \mathbb{R}^{16 \times 16 \times 3}, i \in \{1, 2, ... , 196\}$. Following the Masked Auto-Encoder (MAE)~\cite{he2022mae}, we randomly mask $75\%$ of these tokens and only feed the resting $25\%$ into the following networks. A convolution layer with three kernels $16 \times 16$ is used to project the image patches $P_i$ into the token embeddings $P^j_{emb} \in \mathbb{R}^{1 \times 768}, j \in \{1, 2, ..., 49\}$. The CLS-token is also integrated, therefore, we have the input token embeddings $P_{emb} \in \mathbb{R}^{50 \times 768}$. Meanwhile, we introduce the position encoding $Z^1_{pos} \in \mathbb{R}^{50 \times 768}$ to encode the spatial coordinates of input tokens. Similar to existing works, we random initialize the position encoding and add it with token embeddings, therefore, we have $\tilde{P}_{emb} = Z^1_{pos} + P_{emb}$.

After the input embedding $\tilde{P}_{emb} \in \mathbb{R}^{50 \times 768}$ is obtained, we feed them into ViT-B/16~\cite{dosovitskiy2020image} encoder which contains 12 Transformer blocks. Each Transformer block consists of layer normalization, multi-head self-attention (MSA), and Multi-Layer Perceptron (MLP), as shown in the right part of Fig.~\ref{framework}. The output $\bar{P} \in \mathbb{R}^{50 \times 768}$ from the Transformer encoder is the same as the input tokens. After a 512-dimensional linear projection layer is used to project the output $\bar{P} \in \mathbb{R}^{50 \times 768}$ from Transformer encoder into the decode embedding $\bar{P}^k_{emb} \in \mathbb{R}^{1 \times 512}, k \in \{1, 2, ..., 50\}$. Two kinds of input tokens are fed into the Transformer decoder, i.e., the mask tokens $P_{mask} \in \mathbb{R}^{147 \times 512}$ and the encoded visible tokens $\bar{P}_{emb} \in \mathbb{R}^{50 \times 512}$. Note that, the mask tokens~\cite{kenton2019bert} are shared and learnable vectors. Position encoding $Z^2_{pos} \in \mathbb{R}^{197 \times 512}$ is also introduced and combined with input tokens. The decoder network contains 8 Transformer blocks and is only used in the pre-training phase for image reconstruction.

The mean square error (MSE) over pixel space from the masked token in the original image and the reconstructed token is adopted as the reconstruction loss to optimize the MAE module, which can be written as: 
\begin{equation}
L_{r} = \frac {1} {N_{m}} \sum_{t\in P_{m}} || V_{t} - V^r_{t} ||_{2}
\end{equation}
where $V$ is the RGB pixel value of the input image, and $V^r$ is the predicted pixel value. $N_{m}$ is the number of masked pixels, $P_{m}$ is the index of pixel of the mask, $||*||_2$ refers to the $L_{2}$ loss.

\noindent 
\textbf{Structural Prior Module. }
The aforementioned MAE can already achieve good performance on vehicle-based perception, however, we think this model can be further improved based on the design of auxiliary tasks. The structural prior guided image reconstruction is the first auxiliary task we proposed in this work, as shown in  Fig.~\ref{framework}. The core motivation of this module is that we observed vehicles to be distinct from typical objects, possessing significant contour information such as horizontal and vertical lines, curves, and so on. This structural information about vehicles will play a crucial role in effectively enhancing vehicle image reconstruction.

In our practical implementation, we first adopt the edge detector BDCN (Bi-Directional Cascade Network)~\cite{he2020bdcn} to obtain the outline of the vehicle. Then, similar to the operations conducted on the input vehicle image, the skeleton map is also partitioned into non-overlapping patches, and each patch is projected into a token whose scale is $196 \times 768$ using a convolution layer. Differently, we don't conduct masking operations on the skeleton map.The CLS-token is also integrated, accordingly, we initialize the position encoding $Z^3_{pos} \in \mathbb{R}^{197 \times 768}$ and add it with the tokens as the input of the Transformer encoder. Note that this encoder is shared with the Transformer encoder used for vehicle image encoding. The skeleton feature vectors are treated as a guide for vehicle reconstruction.

To achieve structural priori-guided masked token reconstruction, we adopt the widely used knowledge distill~\cite{bao2021beit} scheme. We project the skeleton tokens and reconstructed invisible tokens into probability distributions with K dimensions (i.e., $P^{patch}_{\theta^{\prime}}(F^s_{i})$ and $P^{patch}_{\theta} (F^t_{i})$ in Eq.~\ref{LmimLoss}, respectively) using two separate projection layers whose parameters are $\theta^{\prime}$ and $\theta$. $F^{s}$ is the skeleton feature output by the structural prior model. $F^t$ is the feature corresponding to the reconstructed invisible token. The distill loss function can be formally written as: 
\begin{equation}
\label{LmimLoss}
L_{mim} = -\sum^N_{i=1} P^{patch}_{\theta^{\prime}}(F^s_{i})^Tlog P^{patch}_{\theta} (F^t_{i}) 
\end{equation} 
where $N$ is the number of masked patches in the encoding phase.  
In order to obtain better visual semantic information, we also project the skeleton CLS token and reconstructed CLS token to obtain their respective classification distributions. The distill loss function can be formally written as:
\begin{equation}
\label{LclsLoss}
L_{cls} =  -P^{cls}_{\theta^{\prime}}(F^s)^Tlog P^{cls}_{\theta} (F^t) 
\end{equation}




\noindent 
\textbf{Semantic Prior Module. } 
The MAE focuses on reconstructing the vehicle image in an auto-regression manner, and our newly proposed structural prior helps the reconstruction from the spatial contour layout. Although better performance can be obtained, however, the high-level semantic information about vehicles is still ignored. For example, on automotive manufacturer websites, you can find pervasive descriptive introductions, attribute information, specific parameters, and hardware configurations about vehicles. Multi-modal pre-training can leverage more clues and  this allows the big model to better understand the vehicle.

In this work, we adopt pre-trained vision-language model CLIP to process the vehicle image $I$ and corresponding natural language descriptions $T = [w_1, w_2, ..., w_m]$, where $w_i$ denotes the $i^{th}$ English word. Specifically, the language encoder of CLIP~\cite{radford2021learning} embeds the sentence into a set of word tokens $W \in \mathbb{R}^{12693 \times 512}$. Meanwhile, the vision encoder of CLIP~\cite{radford2021learning} transforms the vehicle image into visual tokens $V^c \in \mathbb{R}^{1 \times 512}$. Note that, the parameters of both  CLIP~\cite{radford2021learning} visual and language encoder are fixed. The cosine similarity between the 
CLIP~\cite{radford2021learning} visual features and MAE Transformer decoded features are considered in the reconstruction process. In addition, we also introduce cross-modality contrastive learning to achieve semantic-aware feature learning in the decoding phase.

We normalize the CLIP visual features and MAE transform decode features using L2 normalization and calculate the similarity loss between the two features as follows:
\begin{equation}
L_{cf} = (\frac {F^t} {||F^t||_{2}}- \frac {V^c} {||V^c||_{2}})^2  = (\widetilde{F^t} - \widetilde{V^c} )^2
\end{equation}
where $F^t$ is decoded features from MAE, and $V^c$ is CLIP visual features.

On the other hand, we also consider the consistency constraint between the similarity of CLIP text-visual features and CLIP text-MAE decoded visual features. To be specific, the similarity between the text $W=w_{j}, j=\{1, 2, ..., m\}$ and the MAE decoded features is firstly calculated by: 
\begin{equation}
s_{j}(\widetilde{F^t},W) = \frac {exp((\widetilde{F^t}*w_{j})/ \tau)} {\sum^M_{m=1} exp((\widetilde{F^t}*w_{m})/ \tau)}
\end{equation} 
where $\tau$ is a temperature hyper-parameter and we set it as 1 in our experiments. 
The similarity distribution between the text and MAE decoded features can be obtained via:
\begin{equation}
S(\widetilde{F^t},W) = [s_{1}(\widetilde{F^t},W),s_{2}(\widetilde{F^t},W),...,s_{M}(\widetilde{F^t},W)]. 
\end{equation}
Similarly, we can get the similarity distribution between the text and CLIP visual features via: 
\begin{equation}
S(\widetilde{V^c},W) = [s_{1}(\widetilde{V^c},W),s_{2}(\widetilde{V^c},W),...,s_{M}(\widetilde{V^c},W)]
\end{equation} 
The consistency constraint can be achieved by minimizing the two similarity distributions, i.e., 
$KL(S(\widetilde{V^c}, W), S(\widetilde{F^t}, W))$, here, $KL$ is short for relative entropy loss function. 
The regularization term of the similarity distribution between the MAE decoded features and CLIP text features is also introduced to enhance our model, which can be written as $H(S(\widetilde{F^t}, W))$, here, $H$ is short for entropy.
Therefore, the regularized similarity distribution consistency loss function can be formulated as: 
\begin{equation}
L_{cs} = \sum^N_{i=1} KL(S(\widetilde{V^c},W),S(\widetilde{F^t},W)) + \sum^N_{i=1} H(S(\widetilde{F^t},W))
\end{equation}

Finally, the overall loss function of our proposed VehicleMAE can be expressed as:
\begin{equation}
L = L_{r} + L_{mim} +L_{cls} + L_{cf} + L_{cs}. 
\end{equation}

\begin{figure}[!htp]
\center
\includegraphics[width=3.3in]{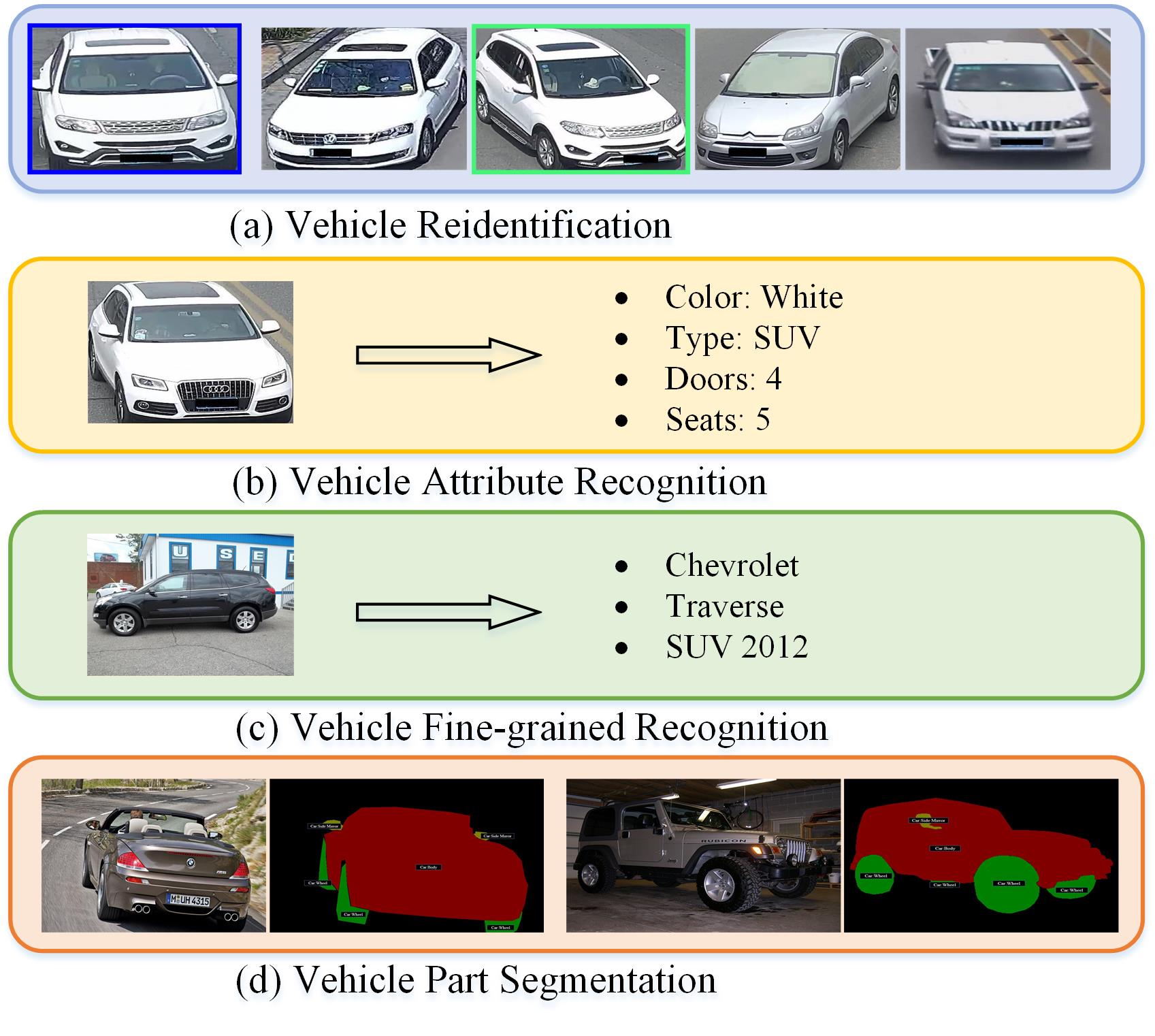}
\caption{Illustration of the downstream tasks evaluated in this work, including (a) vehicle re-identification, (b) attribute recognition, (c) fine-grained recognition, and (d) part segmentation.}    
\label{downstreamTasks}
\end{figure}

\subsection{Downstream Tasks} 

In this work, four downstream tasks are adopted to validate the effectiveness and generalization of our proposed VehicleMAE big model, including vehicle re-identification, attribute recognition, fine-grained recognition, and part segmentation. A brief introduction to these tasks can be found below. 

\noindent 
\textbf{Vehicle Re-identification. } 
Given one query vehicle image, the target of vehicle re-identification is to find the matched vehicles from a set of candidate images that share the same ID with the query image. An illustration of query vehicle image and searched images are given in Fig.~\ref{downstreamTasks} (a). In this work, we validate our VehicleMAE based on TransReID~\cite{he2021transreid}.

\begin{figure}
\center
\includegraphics[width=3.3in]{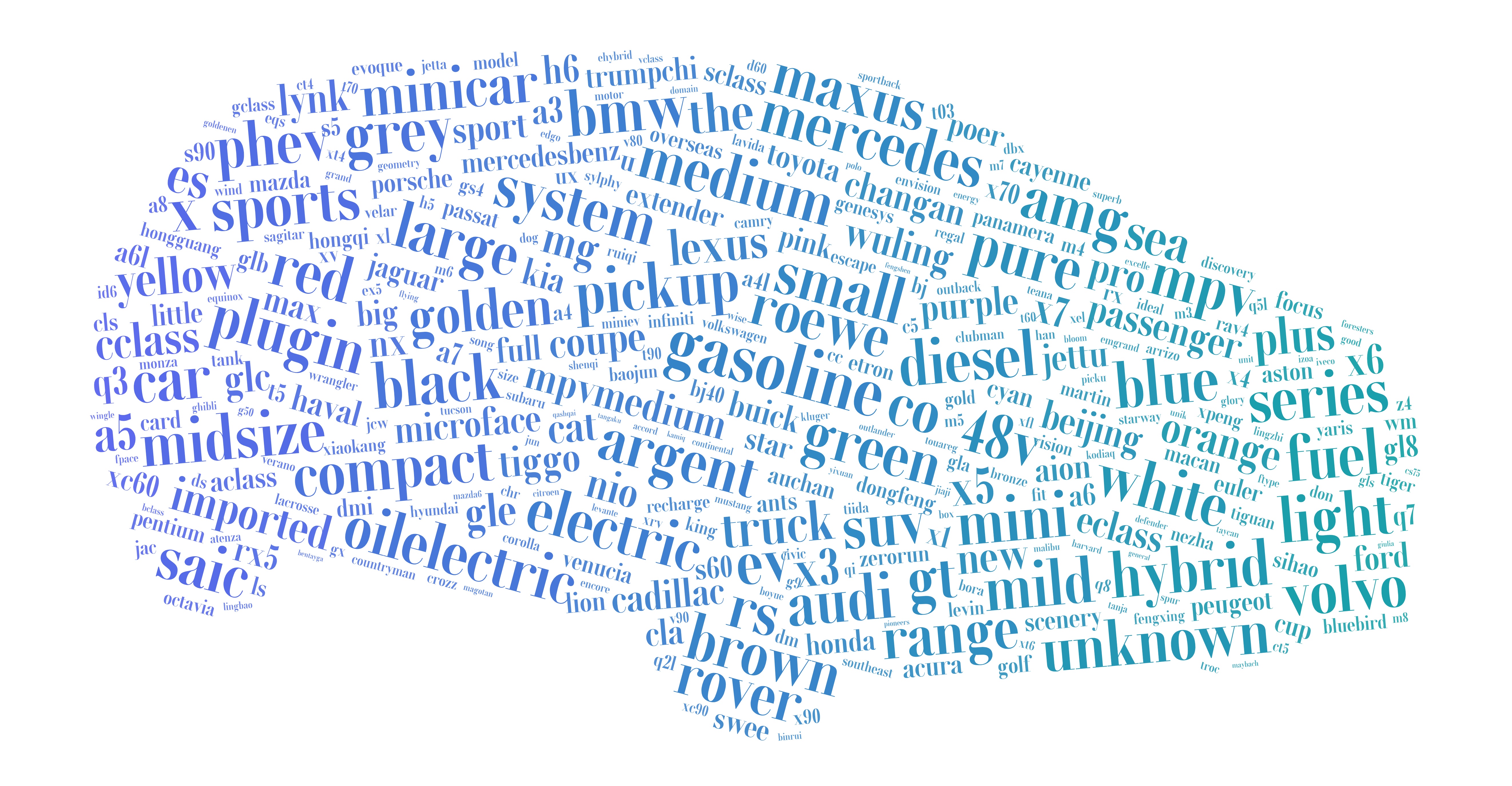}
\caption{Tag clouds of text descriptions in our Autobot1M dataset.}    
\label{word_frequency}
\end{figure}

\noindent 
\textbf{Vehicle Attribute Recognition. } 
This task aims at describing the attributes of given the vehicle image, such as the brand, color, type, etc. It can be seen as a multi-label classification or multi-task learning problem. In this work, we validate our VehicleMAE based on VTB~\cite{cheng2022VTB} which is developed for pedestrian attribute recognition.

\noindent 
\textbf{Vehicle Fine-grained Recognition. } 
Different from the standard recognition problem, the fine-grained vehicle recognition targets capture detailed clues to discriminate different vehicles. In this work, we validate our VehicleMAE based on TransFG~\cite{he2022transfg}.

\noindent 
\textbf{Vehicle Part Segmentation. } 
The segmentation task belongs to the pixel-level perception problem, which targets classifying each pixel into a set of pre-defined labels. An example of vehicle part segmentation is illustrated in Fig.~\ref{downstreamTasks} (d). In this work, we validate our VehicleMAE based on SETR~\cite{zheng2021rethinking}, which is developed for semantic segmentation.

\section{Experiments}  \label{experiments}

\begin{figure*}[!htp]
\center
\includegraphics[width=7in]{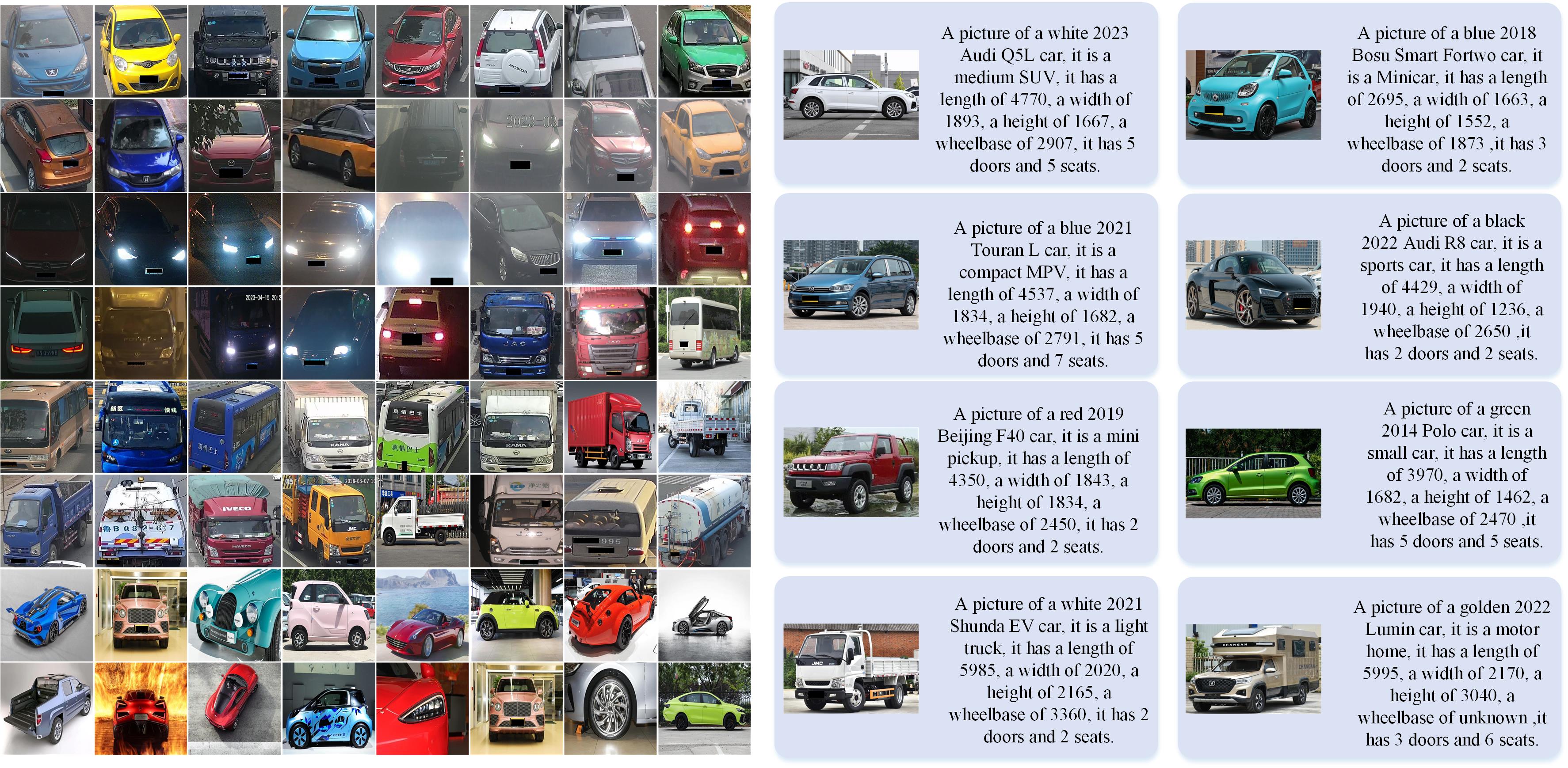}
\caption{Representative samples of our proposed Autobot1M dataset.} 
\label{dataSamples}
\end{figure*}

\subsection{Datasets} \label{DSdatasets}  

In this work, the proposed VehicleMAE big model is pre-trained on our newly proposed \textbf{Autobot1M} dataset. Then, we validate its effectiveness and generalization on three datasets corresponding to four downstream tasks. A brief introduction to these datasets is given below.

\noindent 
$\bullet$ \textbf{Pre-training Dataset.} \label{autobot1m}
In this paper, we propose a large-scale, high-quality vehicle-centric dataset, termed Autobot1M. It contains 1026394 vehicle images from diverse scenarios and sources, including existing vehicle dataset CompCars~\cite{yang2015large} and VERI-Wild ~\cite{lou2019veri}. There are 732112 surveillance images and 294282 network images. These images fully reflect the key features of vehicles, such as illumination, motion blur, viewpoints, and occlusion. Our dataset also considers multiple challenging factors such as illumination, motion blur, viewpoints, occlusion, etc. It is worth noting that part of these images are crawled from the Internet and the corresponding natural language descriptions are available for the pre-training. Some representative samples of our data can be found in Fig.~\ref{word_frequency} and Fig.~\ref{dataSamples}.

\noindent 
$\bullet$ \textbf{Downstream Datasets. }  
In our four different downstream tasks, three datasets are adopted for the downstream validation, including the \textbf{VeRi} dataset~\cite{liu2016Veri776}, \textbf{Stanford Cars} dataset~\cite{krause20133Stanfordcars}, and \textbf{PartImageNet} dataset~\cite{he2022partimagenet}.

\noindent 
\textbf{VeRi} dataset~\cite{liu2016Veri776}~\footnote{\url{https://paperswithcode.com/dataset/veri-776}} is used for the vehicle attribute recognition and vehicle re-identification task. It contains over 50,000 images of 776 vehicles captured by 20 cameras. These images are collected from practical scenarios. The authors provide varied attributes, such as the BBoxes, types, colors, and brands, etc. These images also reflect key challenges for vehicle-based perception, such as various viewpoints, illuminations, resolutions, and occlusions. In Vehicle Re-identification experiments, we split these images into the training and testing subset which contains 37715 and 11579 samples. In Vehicle Attribute Recognition experiments, we split these images into the training and testing subset which contains 37746 and 11579 samples.

\noindent
\textbf{Stanford Cars}  dataset~\cite{krause20133Stanfordcars}~\footnote{\url{https://www.kaggle.com/datasets/jessicali9530/stanford-cars-dataset}} is adopted for the training and evaluation of fine-grained vehicle classification. It contains 16,185 images which correspond to 196 classes of vehicles. The categories of this dataset reflect the Make, Model, and Year of vehicles. The training and testing subset contains 8,144 and 8,041 images respectively. The resolution of the images in this dataset is $360 \times 240$.

\noindent 
\textbf{PartImageNet} dataset~\cite{he2022partimagenet} \footnote{\url{https://github.com/TACJu/PartImageNet}} is utilized for the evaluation of vehicle part segmentation. High-quality part segmentation annotations are provided. It contains about 24000 images which correspond to 158 classes from ImageNet. 11 super-categories can be grouped based on these classes. Following existing works~\cite{zheng2021rethinking}, we split this dataset into the training and testing subset which contains 1657 and 381 samples, respectively.

\begin{table*}[!htp]
\centering 
\small 
\caption{Experimental results of ours and other pre-trained models on vehicle attribute recognition (VAR), re-identification (V-Reid), fine-grained recognition (VFR), and partial segmentation (VPS).} 
\label{table1} 
\vspace{5pt} 
\begin{tabular}{cc|ccccc|cc|c|cc}  
\hline \toprule [0.5 pt]
\multicolumn{2}{c}{ }
&\multicolumn{5}{c}{\textbf{VAR}} &\multicolumn{2}{c}{\textbf{V-Reid}}
&\multicolumn{1}{c}{\textbf{VFR}}
&\multicolumn{2}{c}{\textbf{VPS}} \\
\textbf{Method}  &\textbf{Dataset} &\textbf{mA} &\textbf{Accuracy} & \textbf{Precision} & \textbf{Recall} & \textbf{F1} &\textbf{mAP}& \textbf{R1} &\textbf{Accuracy} & \textbf{mIou} & \textbf{mAcc}   \\ 
\hline 
Scratch   &- &84.67 &80.86 &84.66 &85.77 &84.90    &35.3 &57.3   &24.8   &49.36 &59.22 \\
MoCov3  &ImagNet-1K &90.38 &93.88 &95.57 &95.48 &95.33   &75.5 &94.4   &91.3  &73.17 &78.60 \\
DINO  &ImagNet-1K &89.92 &91.09 &92.84 &93.60 &93.11   &64.3 &91.5   & -  &68.43 &73.37 \\
IBOT  &ImagNet-1K &89.51 &90.17 &91.95 &93.03 &92.37   &68.9 &92.6   &81.1  &66.03 &71.06 \\
\hline 
MAE  &ImagNet-1K &89.69 &93.60 &94.81 &95.54 &95.08   &76.7 &95.8   &91.2  &69.54 &75.36 \\
MAE  &Autobot1M &90.19 &94.06 &95.45 &95.68 &95.43   &75.5 &95.4   &91.3  &69.00 &75.36 \\
VehicleMAE  &Autobot1M &92.21 &94.91 &96.00 &96.50 &96.17  &85.6 &97.9   &94.5   &73.29 &80.22 \\
\hline \toprule [0.5 pt]
\end{tabular}
\end{table*}

\subsection{Evaluation Metric}
In this work, multiple evaluation metrics are used for different downstream tasks, including mA, Accuracy (Acc), Precision, Recall, F1-score, map, R1, mIoU, and mAcc. 

To be specific, the formulaic expression of $Precision$, $AP$, $Recall$, $IoU$, and $Accuracy$ can be written as: 
\begin{flalign}  
& Precision = TP/(TP + FP)  \\
& AP = \frac{1}{M} \sum_{j=1}^{M}(Precision_{j} ) \\
& Recall = TP/(TP+FN)   \\
& IoU = TP/(TP+FN+FP)   \\
& Accuracy = (TP+TN)/(TP+TN+FN+FP) 
\end{flalign} 
where $TP$, $N$, $FP$, and $FN$ represent the number of true positives, true negatives, false positives, and false negatives, respectively. $M$ is the number of images for a category target. As a general evaluation metric, the F1-score considers both Precision and Recall metrics and can be formulated as:
\begin{equation}
F1-score = 2 * \frac {Precision * Recall} {Precision + Recall}
\end{equation}

The label-based evaluation metric mean Accuracy (mA) is defined as: 
\begin{equation}
mA = \frac {1} {2N} \sum_{i=1}^{N}( \frac {TP_{i}} {TP_{i} + FN_{i}} + \frac {TN_{i}} {TN_{i} + FP_{i}})
\end{equation}
where $N$ is the number of attributes, $TP_{i}$ and $TN_{i}$ are the number of true positives, true negatives of the $i$-th attribute, and $FN_{i}$ and $FP_{i}$ are the number of false positives, false negatives of the $i$-th attribute. 
Rank-1 (R1) is an important metric in vehicle re-identification, the task takes the remaining images and the query image and calculates the similarity, while Rank-k indicates the accuracy of the presence of the top-$k$ images with the highest similarity having the same ID as the query image. 
The mAP (mean Average Precision), mIoU, and mAcc denote the mean value of the corresponding metric.

The  (mAP) is defined as:
\begin{equation}
mAP = \frac {\sum_{k=1}^{C_{1}}(AP_{k} )} {C_{1}}
\end{equation}
where $C_{1}$ is the number of categories for the target. 
mIoU is a widely used metric for the semantic segmentation task which can be formulated as: 
\begin{equation}
mIoU = \frac {\sum_{k=1}^{C_{2}}(IoU_{k} )} {C_{2}}
\end{equation}
where $C_{2}$ is the number of categories for the target. 
mAcc is a popular metric for the fine-grained identification task, which can be written as: 
\begin{equation}
mAcc = \frac {\sum_{k=1}^{C_{3}}(Accuracy_{k} )} {C_{3}}
\end{equation}

\subsection{Implementation Details}
In our pre-training phase, the learning rate is set as 0.00025, and the weight decay is 0.04. The AdamW~\cite{loshchilov2018decoupled} is selected as the optimizer to train our model. The batch size is 512 and training for a total of 100 epochs on our Autobot1M dataset. The tradeoff parameters between various loss functions are set as 4, 0.02,0.02, 2, and 0.1, respectively. All the experiments are implemented using Python based on the deep learning toolkit PyTorch~\cite{paszke2019pytorch}. A server with four RTX3090 GPUs is used for the pre-training. About 58 hours are needed for our pre-training phase.

In the finetuning phase, we conduct different settings for the four downstream tasks. To be specific, for the vehicle re-identification task, we finetune the pre-trained model using a learning rate of 0.004. A total of 120 epochs are trained with batch size 32. We follow the same settings of TransReID~\cite{he2021transreid} for other parameters. For the vehicle attribute recognition task, the learning rate and batch size are set as 0.0004 and 32. A total of 40 epochs are trained with the AdamW optimizer. For the vehicle fine-grained recognition task, we set the learning rate as 0.02 and the batch size as 8. The other parameters shared with the TransFG~\cite{he2022transfg} and a total of 320 epochs are optimized. For the vehicle part segmentation task, we finetune the pre-trained model using a learning rate of 0.004, batch size of 2, and a total of 80000 iterations for this experiment.


\subsection{Compare with Other SOTA Algorithms}   \label{compSOTA}

In this experiment, four downstream tasks are used for the validation of our VehicleMAE pre-trained big model. Three different settings of the training data in the pre-training phase are evaluated, i.e., full data, $20\%$, and $10\%$ of the training data. We compare with our baselines and also other state-of-the-art models on each downstream task, as shown in Table~\ref{table1}. More in detail, the models without the pre-training (i.e., learning from scratch), the pre-trained MAE model on the ImageNet dataset, and the pre-trained MAE model on our newly proposed Autobot1M dataset.

For vehicle attribute recognition, we report and compare the attribute recognition results on the Veri-776 dataset. The baseline approach VTB~\cite{cheng2022VTB} proposed by Cheng et al. in the year 2022 is selected and equipped with the pre-trained big models to validate the effectiveness. We can find that the VTB achieves $84.67\%$, $80.86\%$, $84.66\%$, $85.77\%$, and $84.90\%$ on mA, Accuracy, Precision, Recall, and F1 metrics, respectively, based on ViT-base when learning from scratch. When initializing the ViT-base backbone network using parameters learned by MAE on the ImageNet dataset, the recognition results can be improved to $89.69\%$, $93.60\%$, $94.81\%$, $95.54\%$, $95.08\%$. This experiment demonstrates that the visual features learned by self-supervised pre-training contribute significantly. When replacing the ImageNet using our proposed Autobot1M dataset, the recognition performance can be further improved which demonstrates that the pre-training on vehicle images performs better than the generalized data in natural scenarios. 
Note that, our proposed pre-trained framework performs the best when compared with the aforementioned models, i.e., $92.21\%$, $94.91\%$, $96.00\%$, $96.50\%$, $96.17\%$. In contrast, existing pre-trained big models like the MoCov3~\cite{MoCov32021}, DINO~\cite{caron2021emerging}, and IBOT~\cite{zhou2021image}, are all inferior to our model. These experimental results and comparisons fully validated the effectiveness of our proposed pre-training strategy for vehicle-based pre-training. Similar conclusions can also be drawn from the other three downstream tasks.

\begin{table*} 
\centering 
\small 
\caption{Ablation study on loss functions in Structural Prior and Semantic Prior.}  
\label{AblationStudy1}  
\vspace{5pt} 
\begin{tabular}{c|cc|cc|ccccc|cc|c|cc} 
\hline  \toprule [0.5 pt] 
\textbf{MAE Loss} &\multicolumn{2}{c|}{\textbf{Structural Prior}}  &\multicolumn{2}{c|}{\textbf{Semantic Prior}} &\multicolumn{5}{c|}{\textbf{VAR}} &\multicolumn{2}{c|}{\textbf{V-ReID}} &\textbf{VFR} &\multicolumn{2}{c}{\textbf{VPS}}\\
$L_{r}$  &$L_{mim}$ & $L_{cls}$    &$L_{cs}$ & $L_{cf}$& mA & Acc & Prec & Rec & F1 & mAP &R1 &Acc &mIou &mAcc \\ 
\hline 
\checkmark& & & & &90.19 &94.06 &95.45 &95.68 &95.43 &75.5 &95.4 &91.3 &69.00 &75.36 \\
\checkmark&\checkmark & & & &91.27 &94.11 &95.29 &95.82 &95.50 &79.7 &96.1 &93.2 &70.34 &75.70 \\
\checkmark&\checkmark &\checkmark & & &91.71 &94.54 &95.65 &96.28 &95.88 &83.4 &96.6 &93.7 &70.65 &76.04 \\
\hline 
\checkmark& & &\checkmark & &92.12 &94.28 &95.42 &96.23 &95.71 &84.1 &97.1 &94.1 &71.90 &76.47 \\
\checkmark& & &\checkmark &\checkmark &92.15 &94.58 &95.69 &96.36 &95.92 &85.2 &97.1 &94.3 &71.87 &77.93 \\
\checkmark&\checkmark &\checkmark &\checkmark &\checkmark &92.21 &94.91 &96.00 &96.50 &96.17 &85.6 &97.9 &94.5 &73.29 &80.22 \\
\hline  \toprule [0.5 pt] 
\end{tabular}
\end{table*}

\begin{table*}
\centering 
\caption{Ablation study on the ratio of masked tokens.} 
\label{ratioMaskResults} 
\vspace{5pt} 
\begin{tabular}{c|ccccc|cc|c|cc} 
\hline  \toprule [0.5 pt] 
\multirow{2}{*}{\raggedright \textbf{Ratio of Masked Token}}  &\multicolumn{5}{c|}{\textbf{VAR}} &\multicolumn{2}{c|}{\textbf{V-ReID}} &\textbf{VFR} &\multicolumn{2}{c}{\textbf{VPS}}\\
&mA & Acc & Prec & Rec & F1 & map &R1 &Acc &mIoU &mAcc \\ 
\hline 
0.25 &90.48 &94.34 &95.63 &96.02 &95.72 &84.9 &97.0 &94.0 &72.31 &78.20 \\
0.50 &91.88 &94.35 &95.55 &96.11 &95.72 &85.2 &97.3 &94.3 &71.90 &77.66 \\
0.75 &92.21 &94.91 &96.00 &96.50 &96.17 &85.6 &97.9 &94.5 &73.29 &80.22\\
0.85 &90.73 &94.18 &95.32 &95.97 &95.55 &82.1 &96.3 &93.5 &70.91 &77.12 \\
\hline  \toprule [0.5 pt] 
\end{tabular}
\end{table*}

\begin{table*} 
\centering
\small 
\caption{Results of different scales of training data used in downstream tasks. - denotes no corresponding results.} 
\vspace{5pt}
\label{DifferentDataVolumesResults} 
\begin{tabular}{c|c|c|ccccc|cc|c|cc} 
\hline  \toprule [0.5 pt] 
\multirow{2}{*}{\raggedright \textbf{Training Data}} &\multirow{2}{*}{\raggedright \textbf{Method}}&\multirow{2}{*}{\raggedright \textbf{Dataset}} &\multicolumn{5}{c|}{\textbf{VAR}} &\multicolumn{2}{c|}{\textbf{V-ReID}} &\multicolumn{1}{c|}{\textbf{VFR}} &\multicolumn{2}{c}{\textbf{VPS}}\\
 & & & mA & Acc & Prec & Rec & F1 & mAP &R1 &Acc &mIoU &mAcc \\ 
\hline 
\multirow{4}{*}{\raggedright $20\%$} 
 &Scratch &- &80.94 &71.33 &76.18 &79.64 &77.27 &25.2 &34.9 &7.1 &39.87 &49.50 \\
 &MAE &ImageNet-1K &89.32 &92.65 &94.35 &94.87 &94.41 &64.8 &89.7 &42.5 &64.86 &72.20 \\
 &MAE &Autobot1M &89.58 &92.36 &94.09 &95.06 &94.29 &60.0 &85.5 &66.5 &65.04 &70.81 \\
 &VehicleMAE &Autobot1M &91.50 &94.53 &95.74 &96.33 &95.91 &80.9 &95.2 &83.58 &68.72 &76.02   \\
 \hline
\multirow{4}{*}{\raggedright $10\%$}
 &Scratch &- &78.47 &66.48 &72.35 &75.47 &73.25 &- &- & 4.5                 &35.44 &46.22 \\
 &MAE &ImageNet-1K &88.61 &90.78 &92.74 &93.64 &92.95 &- &- &17.1           &52.31 &62.30 \\
 &MAE &Autobot1M &86.49 &89.59 &91.61 &93.33 &92.13 &- &- &21.4             &62.35 &69.56 \\
 &VehicleMAE &Autobot1M &89.29& 93.76&94.94 &95.86 &95.25 &- &- & 71.4      &65.09 &71.19 \\
\hline  \toprule [0.5 pt] 
\end{tabular}
\end{table*}

\subsection{Ablation Study} \label{ablationStudy}

\noindent 
\textbf{Effects of Structural Prior.~} 
In this paper, we introduce the structural prior to guide the reconstruction of given vehicle images. Two loss functions are involved here, i.e., $L_{mim}$ and $L_{cls}$. As shown in Table~\ref{AblationStudy1}, we introduce the two loss functions into the pre-training, and the performance is all improved on four downstream tasks. For example, the results are boosted to $91.27\%$, $94.11\%$, $95.29\%$, $95.82\%$, $95.50\%$ when the $L_{mim}$ is adopted, and to $91.71\%$, $94.54\%$, $95.65\%$, $96.28\%$, $95.88\%$ when both $L_{mim}$ and $L_{cls}$ are used. These experimental results and comparison fully validated the effectiveness of our proposed structural prior. Similar conclusions can also be drawn from other tasks.

\noindent
\textbf{Effects of Semantic Prior.} 
We introduce two loss functions for the semantic prior, i.e., $L_{cs}$ and $L_{cf}$. As shown in Table~\ref{AblationStudy1}, $75.5\%$, $95.4\%$ are obtained on mAP and R1 for the vehicle re-identification task when only the MAE loss function is used. When $L_{cs}$ is utilized, the results can be improved to $84.1\%$, $97.1\%$, which are significant improvements. Note that, these results can be further improved to $85.2\%$, $97.1\%$ when both $L_{cs}$ and $L_{cf}$ are used. 
We can achieve the best performance on four downstream tasks when the four loss functions are all used based on MAE, which fully validates the effectiveness of our proposed structural and semantic prior information for the MAE based vehicle reconstruction.

\noindent 
\textbf{Analysis on Ratio of Masked Tokens.} 
As shown in Table~\ref{ratioMaskResults}, we set different ratios of masked tokens to check their influence. Specifically, 0.25, 0.50, 0.75, 0.85 are tested and the experimental results on the four downstream tasks demonstrate that better performance can be obtained when $75\%$ of the input tokens are randomly masked.

\begin{table}
\centering 
\caption{Comparison of FLOPs, MACs, and Parameters of ours and other pre-trained big models.}  
\label{efficiencyMetric} 
\scriptsize 
\begin{tabularx}{0.48\textwidth}{l|c|c|c|c|c}
\hline \toprule [0.5 pt] 
\textbf{Metric}    &\textbf{MoCov3} &\textbf{DINO} &\textbf{IBOT} &\textbf{MAE} &\textbf{VehicleMAE}   \\ 
\hline 
\textbf{FLOPs (G)}  &18.00 &16.89 &18.52 &9.43 &10.98   \\ 
\textbf{MACs (OPs)} &17.58 &16.88 &18.51 &9.43 &10.98   \\
\textbf{Params (M)} &86.57 &108.87 &96.29 &111.65 &121.62   \\ 
\hline \toprule [0.5 pt] 
\end{tabularx}
\end{table}

\begin{figure*}
\center
\includegraphics[width=7in]{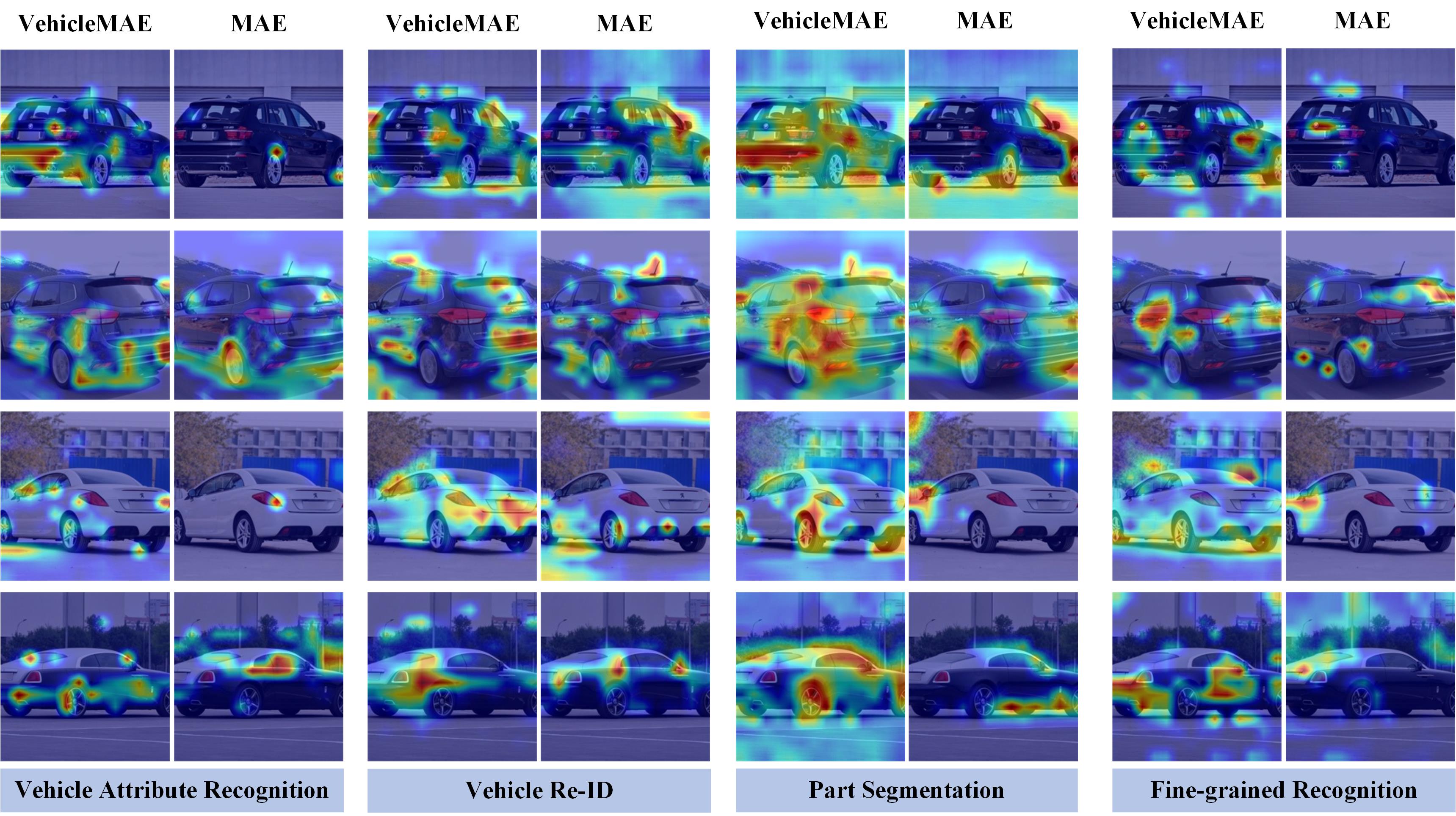}
\caption{Visualization of attention maps on different downstream tasks.}    
\label{attentionmaps}
\end{figure*}

\begin{figure*}[!htp]
\center
\includegraphics[width=7in]{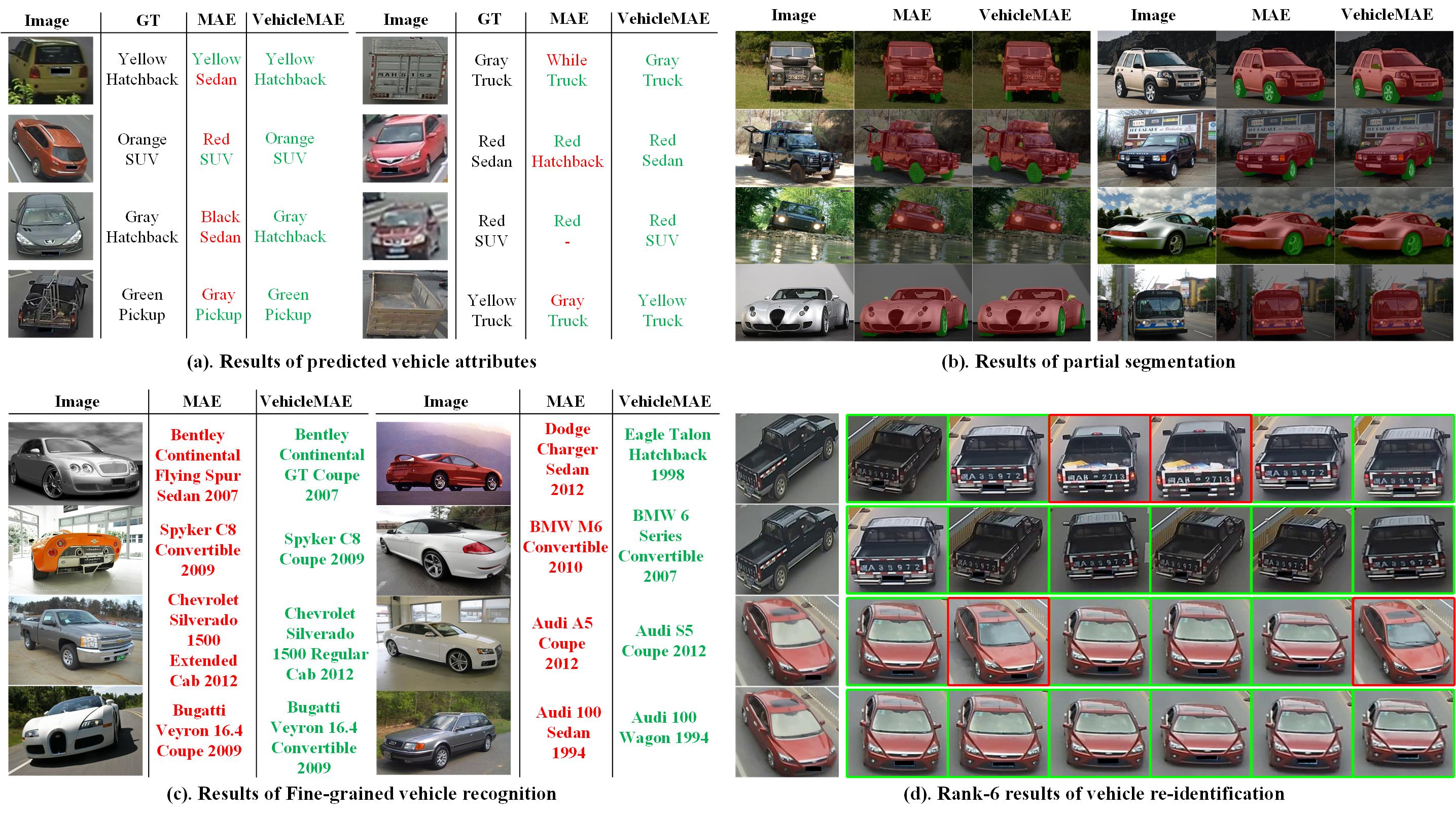}
\caption{Visualization of the experimental results of each downstream task.}    
\label{downstream_vis}
\end{figure*}

\noindent 
\textbf{Analysis on Different Sizes of Training Data in Downstream Tasks.}  
To validate the effectiveness of our pre-trained VehicleMAE on tasks with few training samples, in this part, we utilize $10\%$ and $20\%$ of the training data in the downstream tasks. As shown in Table~\ref{DifferentDataVolumesResults}, we can find that when $10\%$ of training data is used for learning from scratch, the results mIoU and mAcc on vehicle part segmentation are $35.44\%$, $46.22\%$, respectively. When utilizing the pre-trained model using MAE on ImageNet-1K dataset, the results are improved to $52.31\%$, $62.30\%$. This comparison demonstrates that the self-supervised pre-training contributes significantly to the feature representation. If our proposed Autobot1M dataset is used, the results can be further boosted to $62.35\%$, $69.56\%$. We can find that the overall results are best on the part segmentation task when both our dataset and VehicleMAE framework are all adopted, i.e., $65.09\%$, $71.19\%$. 
Similar conclusions can also be drawn from other tasks and the settings of $20\%$ of the training dataset. These experiments fully validated the key contributions of our model and dataset for the vehicle-based perceptron.

\subsection{Efficiency Analysis}  \label{convergeAnalysis} 
As shown in Table~\ref{efficiencyMetric}, we report the FLOPs\footnote{\url{https://pypi.org/project/ptflops/}}, MACs\footnote{\url{https://pypi.org/project/thop/}}, and Parameters of our proposed VehicleMAE and other pre-trained big models to help the readers better understand the efficiency metrics. It is easy to find that the FLOPs and MACs of MAE are all 9.43G, while ours are 10.98G. The parameters of MAE is 111.65M and ours is 121.62M, which demonstrates that the scale of MAE and our proposed VechileMAE are similar. Meanwhile, our model achieves better results on multiple downstream tasks than the MAE framework.

\subsection{Visualization} \label{visualization} 
In this section, we visualize the feature maps of MAE and our VehicleMAE big model, the masked tokens, and reconstructed images. 
As shown in Fig.~\ref{attentionmaps}, we give the feature maps of four vehicle images on four downstream tasks. The GradCAM\footnote{\url{https://mmpretrain.readthedocs.io/en/latest/useful_tools/cam_visualization.html}} is adopted to visualize the attention maps of $11^{th}$ Transformer block. Compared with our baseline MAE, we can find that the heat maps are higher in the key regions. Therefore, our model performs better on the tested vehicle-based perceptron tasks.

\begin{figure}[!htp]
\center
\includegraphics[width=3.3in]{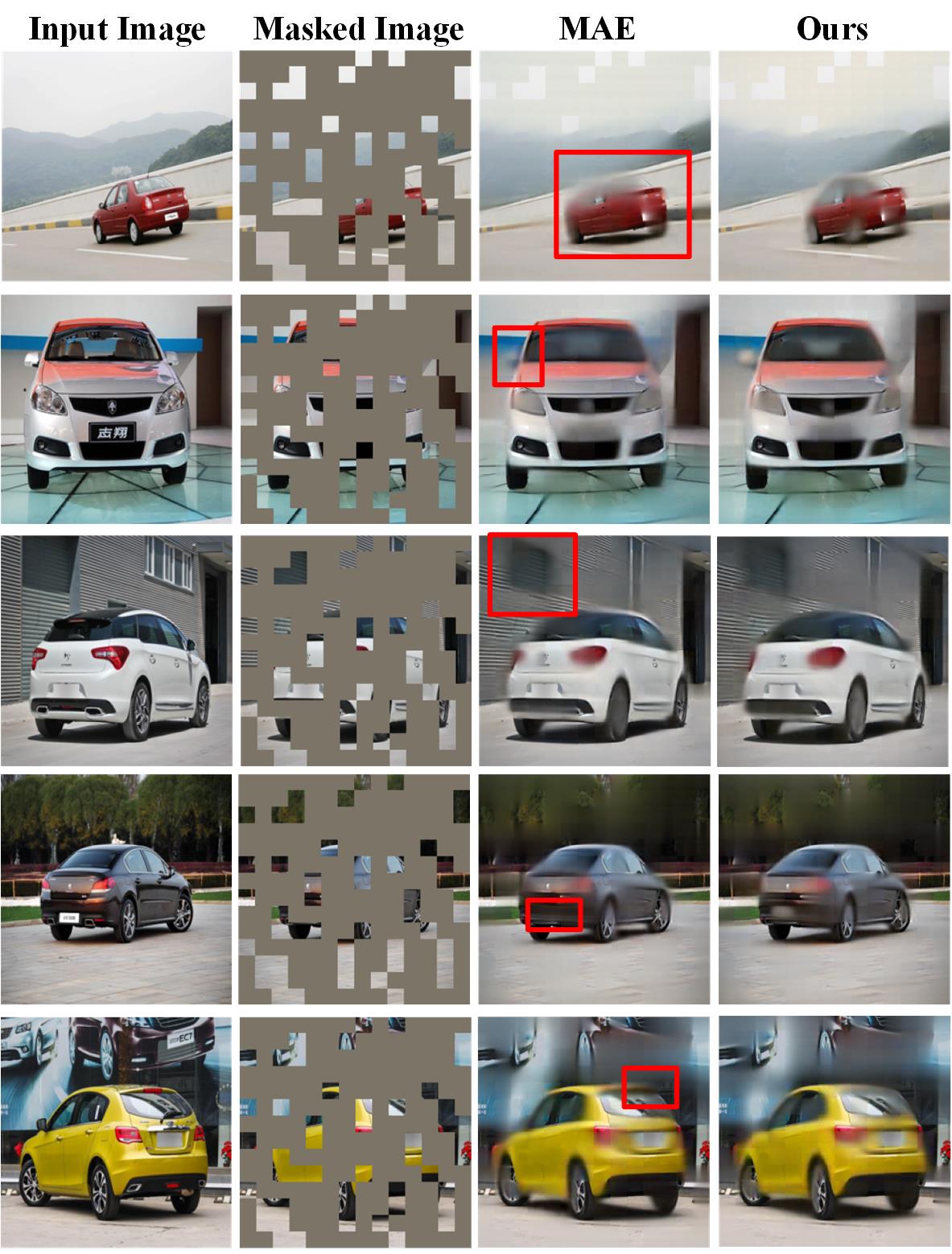}
\caption{Visualization of reconstructed vehicle images using MAE and our newly proposed VehicleMAE.}  
\label{reconst_vis}
\end{figure}

We also visualize the masked tokens in the third and fourth row of Fig.~\ref{reconst_vis}, the first and second columns are input images and images with masked regions. The third and fourth columns are images with reconstructed patches using MAE and our proposed VehicleMAE. Thanks to the structural and semantic prior information, the predicted regions using our model are significantly better than the MAE model, as highlighted in red bounding boxes.

\subsection{Limitation Analysis}  \label{limitationAnalysis} 
According to the aforementioned experiments, we can find that our pre-trained VehicleMAE achieves good performance when our model focuses on the vehicles themselves. However, utilizing our pre-trained VehicleMAE does not lead to a significant improvement in vehicle detection performance, if there are many other background objects present in a large scene. We believe this can be addressed well by introducing more images with vehicles in the pre-training phase. We leave this as our future works.

\section{Conclusion}  \label{Conclusion}
In this paper, we propose the first large-scale pre-trained big model for vehicle-centric perceptron, termed VehicleMAE. Given the vehicle image, we first divide and partition it into non-overlapping patches. Then, we randomly mask these patches with a high ratio (about $75\%$) and project the rest tokens into feature embeddings. The ViT network is adopted as the backbone to process these embeddings, then, masked tokens are padded for reconstruction using a Transformer decoder network. More importantly, the vehicle profile information and high-level natural language descriptions are taken into consideration for effective masked vehicle reconstruction. The sketch lines of vehicles are extracted as a form of structured information to guide vehicle reconstruction. Knowledge distill from the big multimodal model CLIP on the similarity between the paired/unpaired vehicle image-text sample is also considered. To bridge the data gap, we propose a large-scale dataset to pre-train our model termed Autobot1M, which contains about 1M vehicle images and 12693 text information. Four downstream tasks including vehicle attribute recognition, fine-grained recognition, re-identification, and part segmentation, are adopted for the evaluation and comparison. Extensive experiments fully validated the effectiveness and benefits of our VehicleMAE and Autobot1M dataset.

{
    \small
    \bibliographystyle{ieeenat_fullname}
    \bibliography{reference}
}

\end{document}